\documentclass[runningheads]{llncs}

\usepackage{accv}

\usepackage{accvabbrv}

\usepackage{graphicx}
\usepackage{booktabs}
\usepackage{tabularx}
\usepackage[table]{xcolor}

\usepackage[accsupp]{axessibility}  

\usepackage{hyperref}

\usepackage{orcidlink}

\usepackage{listings}
\begin{document}

\title{Drop the Details, Get the Gist: Dynamic Conceptual Representation for Video Recognition}

\titlerunning{Dynamic Conceptual Representation for Video Recognition}

\author{Ryoya Yoshiyama \and Bowen Wang \and Yuta Nakashima}

\authorrunning{R.~Yoshiyama et al.}

\institute{The University of Osaka, Japan\\
\email{\{yoshiyama,wang,n-yuta\}@im.sanken.osaka-u.ac.jp}}

\maketitle

\begin{abstract}
Concept Bottleneck Models (CBMs), which route predictions through a layer of human-interpretable \emph{concepts}, have substantially improved the interpretability of visual classification by exposing an explicit, language-based rationale for each prediction. Most existing work focuses on static images, and temporal dynamics in videos have been underexplored in conceptual representation. We propose a framework that extends concept-based image recognition to procedural video understanding. It represents a multi-step video as a temporal sequence of segment-level concept score vectors and feeds this sequence into a Transformer encoder to predict the class label. We also explore a way to estimate the contribution of each concept for further interpretability. We evaluated three concept sets of different granularities on cooking and household-activity videos and found that coarser-grained concepts are sufficient for classifying these videos, which implies that fine-grained textual descriptions may be unnecessary. 
\end{abstract}

\section{Introduction}
\label{sec:intro}

\begin{figure}[t]
  \centering
  \includegraphics[width=\linewidth]{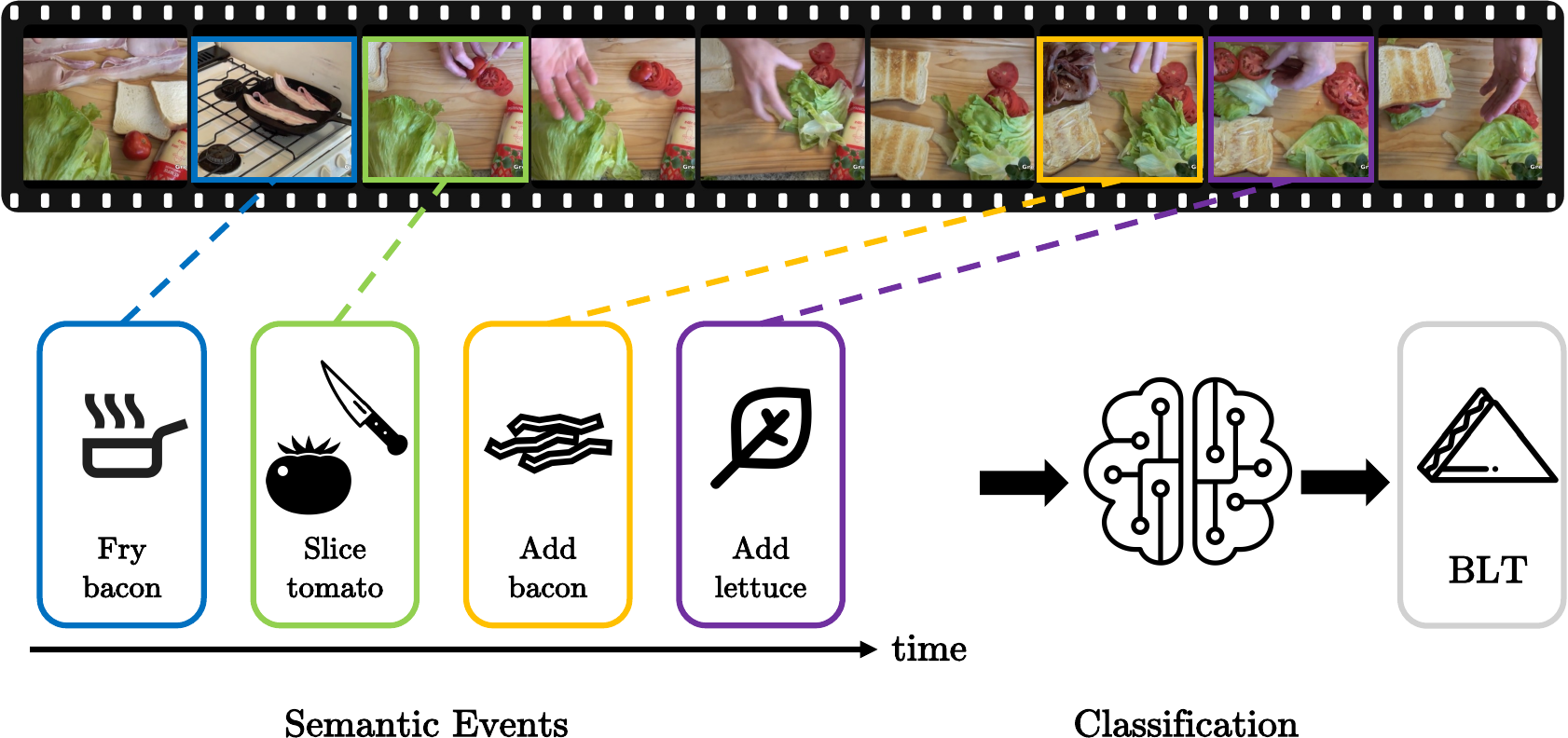}
  \caption{Our approach mirrors human video understanding: a video is parsed into a sequence of semantic events (concepts), and a temporal reasoning module integrates these events to produce an interpretable class prediction.}
  \label{fig:teaser}
\end{figure}

Consider a few minutes of video showing someone making a BLT sandwich (\cref{fig:teaser}): \emph{a tomato being sliced on a cutting board, strips of bacon sizzling in a pan, lettuce leaves placed on a slice of bread, and finally the cooked bacon and sliced tomato stacked on top}. A viewer immediately recognizes what is being prepared, not by processing pixel intensities frame by frame, but by identifying a sequence of meaningful events and reasoning about their order. This suggests that human video understanding is inherently temporal and semantic. We may implicitly parse video into discrete events and integrate them over time to arrive at a holistic interpretation.

This observation motivates a model that reasons over meaningful temporal events rather than treating video as an opaque stream of raw features. Learnable representations have driven rapid improvements in video recognition, yet existing models process video precisely in this opaque fashion \cite{c3d,i3d,slowfast,tsm,video-swin}, making it difficult to understand why a particular prediction was made. This is known as the \emph{black-box} problem. In domains that demand high reliability, such as healthcare, autonomous driving, and security, model interpretability is as essential as predictive accuracy \cite{rudin2019stop}.

To address this issue in the image domain, Concept Bottleneck Models (CBMs) \cite{cbm} have attracted considerable attention. A CBM first predicts a set of human-understandable \emph{concepts} (\eg, ``has wings'' and ``has red feet'' for bird classification) from the input, and then derives the final class prediction from these concepts, thereby making the intermediate representation directly inspectable.

While CBMs have been successful in the static-image domain, extending them to video poses substantial challenges. Unlike images, video carries time as an additional, crucial dimension. Understanding an action or event requires capturing not only the content of individual frames but also how concepts evolve over time. Yet videos vary in length, and modeling long-range dependencies among concepts is considerably more complex than in static images. Extending concept-based interpretability to video remains an open challenge, as the model must account for not only what concepts appear but also when they appear.

Our goal is to build a concept-based video recognition model that captures the temporal dependencies within a video while keeping its predictions grounded in human-interpretable concepts. To this end, we employ a Transformer encoder \cite{attention-is-all-you-need}, which is well-suited to modeling dependencies in sequential data, to process our conceptual representation over time. Specifically, we first identify the set of concepts that describe the video domain well and represent each video segment with these concepts. We then feed the resulting temporal sequence of concept scores into a Transformer, which learns temporal relationships among concepts. This allows the model to base its prediction not merely on which concepts appear in the video as a whole but on when they appear.

To validate this approach, we conduct experiments on two datasets of multi-step procedural videos annotated with textual descriptions. Because concept-based approaches to video recognition remain in an early stage, we treat this as a pilot study that uses textual supervision to investigate what kinds of concepts should be extracted from video, as a step toward understanding what concept representations best support video recognition.

\textbf{Contributions}. We compare three concept sets of differing granularity, ranging from detailed textual descriptions to simple object-and-action concepts, for multi-step procedural video classification. Experiments show that all granularity levels achieve comparable accuracy, suggesting that simple, coarse-grained concepts may be sufficient for procedural video recognition. We employ a framework that represents video as a temporal sequence of concept scores and classifies it with a Transformer encoder. Experiments on YouCook2 and the Household Activities subset of ActivityNet show that our model achieves $89.5\%$ and $93.5\%$ accuracy against black-box baselines of $90.7\%$ and $94.3\%$, respectively, suggesting that interpretability and predictive accuracy can potentially be reconciled.

\section{Related Work}
\label{sec:related}

\subsection{Post-Hoc Explanation Methods}
Methods that localize \emph{where} in the input a model attended to have been studied for image recognition. The most representative approaches rely on gradient information, led by Grad-CAM \cite{grad-cam} and its numerous refinements and variants \cite{grad-cam++,layer-cam,legrad2025}. In video recognition, a common strategy extends such methods along the temporal axis \cite{saliency,swagv2022}. For Transformer-based models, visualizing attention maps to reveal the contribution of specific frames or patches is also widely used \cite{timesformer,staa2025}. Other approaches that do not rely on gradients perturb the input with masks or noise and observe the resulting change in the model's prediction score to identify important regions \cite{zeiler2014visualizing,lime,shap,rise,vitshapley2023}. Video extensions of this idea use spatio-temporal masks to visualize the basis for a prediction \cite{li2021towards,revex2024}.

These methods reduce model explanation to visualizing important regions in the input space. In contrast, we adopt a concept-based perspective. Rather than pointing to a specific spatial region of the video, our model presents what it has recognized as a linguistic \emph{concept}, offering a complementary view of the decision process to that given by visual explanations.

\subsection{Concept-Based Representations of Visual Information}
To address the black-box nature of deep models, prior work has sought to map a model's internal representations onto human-interpretable concepts. Early studies probed a trained model post hoc, either by decomposing its internal representation and associating each component with an image patch \cite{ibd,bau2017network}, or by testing whether a human-defined concept influences its prediction \cite{tcav}. Koh \etal \cite{cbm} instead built this idea into the model architecture, proposing CBMs, which first predict a set of concepts from the input image and then derive the final class solely from those concepts. Because prediction proceeds through the explicit stages $x \to \text{concept} \to y$, the concepts underlying each decision can be inspected directly, and numerous extensions have since been proposed to improve accuracy and practicality \cite{cem,post-hoc-cbm,interactive-cbm,probabilistic-cbm}. To remove the need for the large amounts of concept-level annotation required by conventional CBMs, Label-free CBM \cite{label-free-cbm,labo} constructs a concept bottleneck without additional annotation by leveraging a vision-language model (VLM) such as CLIP \cite{clip} and the similarity it induces between images and text.

Concept-based recognition frameworks are thus becoming increasingly prominent in the static-image domain, and several studies have extended this idea to video. These include approaches based on textual concepts \cite{codex2022,knab2025motif} and those using non-linguistic concept representations \cite{video-concept-discovery,pcbear,dance,hipe,video-tcav}. Among text-based approaches, MoTIF \cite{knab2025motif} models the temporal evolution of concept activations using per-concept self-attention and compares object-only, action-only, and combined concept sets. Our study instead compares concept sets of differing granularity to identify which level of abstraction best supports the classification of multi-step procedural videos containing many temporally ordered sub-events.

\subsection{Video Understanding with Vision-Language Models}
Large-scale vision-language models (VLMs) that embed images and text into a shared feature space have developed rapidly in recent years. CLIP \cite{clip}, a representative model, uses contrastive learning on 400 million image-text pairs to enable zero-shot classification of arbitrary images via natural language, and numerous follow-up studies have since improved its training objective, data scale, and efficiency \cite{align,siglip,openclip-scaling}. Building on this success, similar approaches have been actively explored for video. Methods such as VideoCLIP \cite{videoclip} and the Perception Encoder (PE) \cite{bolya2025perception} learn from video-text pairs to obtain feature representations that capture the temporal structure of video.

We leverage the strong video-text alignment ability of these VLMs not for classification itself but for intermediate concept extraction. Specifically, we compute the similarity between each video segment and each concept using the VLM's video and text encoders, and use these similarities as the input representation to our temporal model (\cref{sec:method}).

\section{Method}
\label{sec:method}

\begin{figure}[tb]
  \centering
  \includegraphics[width=0.9\linewidth]{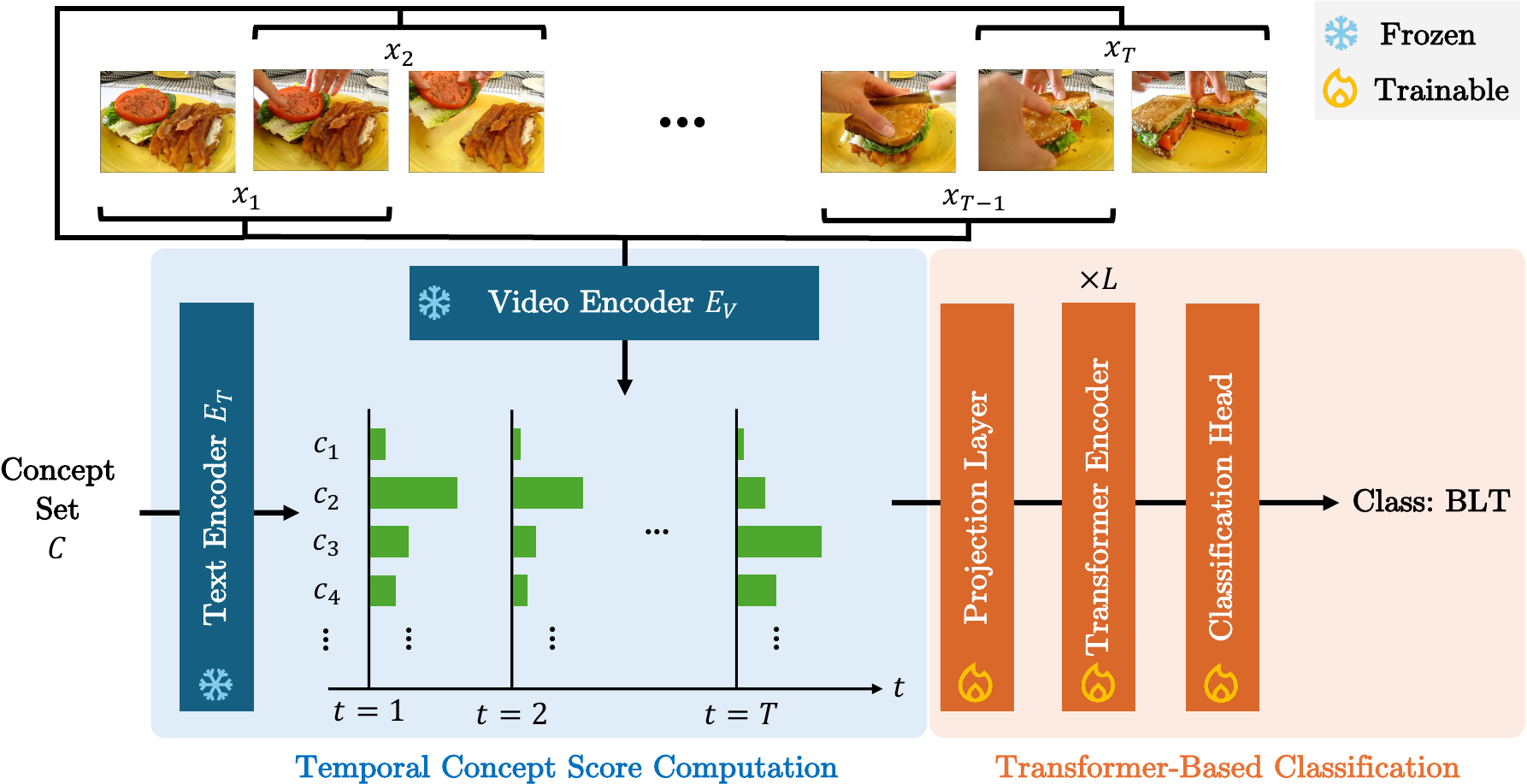}
  \caption{Overview of the proposed framework. Frozen encoders $E_V$ and $E_T$ compute the similarity between each video segment and each concept, and a Transformer trained on the resulting temporal concept scores predicts the class.}
  \label{fig:model-architecture}
\end{figure}

We consider a dataset $\mathcal{D}$ of pairs $(V, y)$, where $V$ denotes an input procedural video and $y$ the class label. As input to our model, a video $V$ is split into a sequence of $T$ overlapping video segments $X = (x_1, x_2, \dots, x_T)$ of a fixed duration. For conceptual representation, we construct a dataset-level concept set $\mathcal{C}$ from the descriptions of training videos. Our method computes similarity between each concept $c \in \mathcal{C}$ and each segment $x_t$ to obtain a concept score vector $s_t$. Finally, a Transformer predicts the class label $y$ from $S = (s_1, \dots, s_T)$. An overview of our method is shown in \cref{fig:model-architecture}.

To determine the concept set $\mathcal{C}$, we use a set of natural-language step descriptions $U = \{u_1, u_2, \dots\}$ of each procedural video $V$ (whose cardinality varies across videos), where the set of descriptions for all videos in the training set of $\mathcal{D}$ is denoted by $\mathcal{U}$. Each description $u \in U$ refers to an action occurring within a specific time interval of the video. We use $\mathcal{U}$ as the source from which concepts are extracted (\cref{sec:method-concepts}).

\subsection{Concept Set Construction}
\label{sec:method-concepts}

The definition of the concept set is a key factor governing both the interpretability and the final accuracy of our model. Describing the diverse events and states in a video in a human-understandable way requires a concept set that is both comprehensive and of an appropriate granularity.

\begin{figure}[tb]
  \centering
  \includegraphics[width=0.9\linewidth]{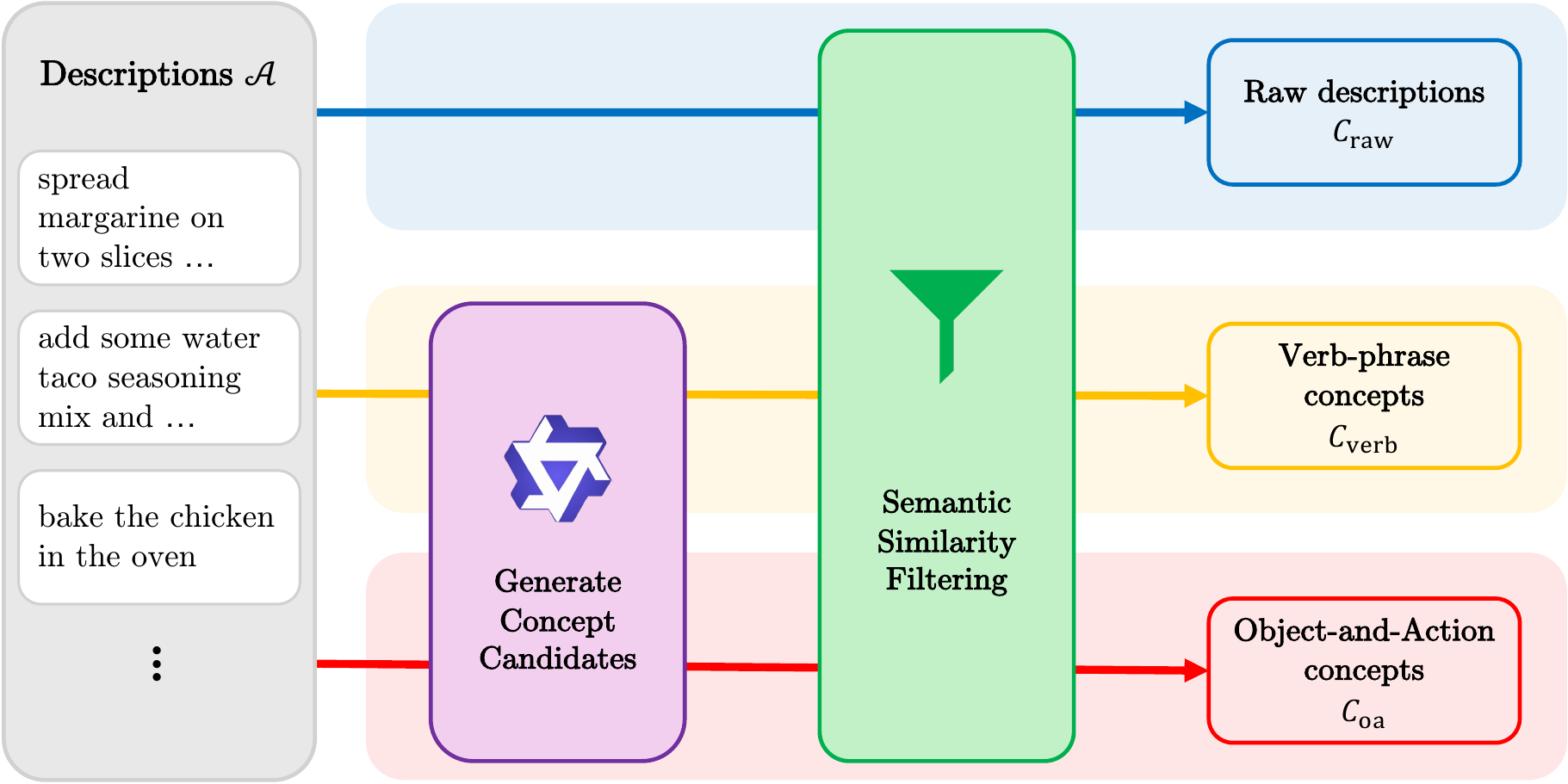}
  \caption{An overview of concept set construction.}
  \label{fig:concept-generation}
\end{figure}

We use the step-level descriptions provided in a procedural video dataset, such as YouCook2 \cite{youcook2}, as the source of concepts. Manually curating a comprehensive concept set from these descriptions would be costly, so we instead automatically generate the concept set with a large language model (LLM). Specifically, we use Qwen3 \cite{qwen3} to extract concepts from $\mathcal{U}$, as illustrated in \cref{fig:concept-generation}. \Cref{tab:concept-examples} shows representative examples of concepts from each set.

To examine how concept granularity affects temporal modeling by the Transformer and the resulting interpretability, we construct three concept sets of different granularities:
\begin{enumerate}
  \item \textbf{Raw descriptions} $\mathcal{C}_\text{raw}$: each natural-language step description is used verbatim as a single concept (\eg, ``add some parmesan cheese and salt to pot''). This setting retains the most information from the original text but, being highly specific, may fail to capture an action shared across different videos.
  \item \textbf{Verb-phrase concepts} $\mathcal{C}_\text{verb}$: the main verb phrases extracted from each step description (\eg, ``add parmesan cheese to pot'', ``add salt to pot''), yielding a medium level of granularity that focuses on the action content important for recognition.
  \item \textbf{Object-and-action concepts} $\mathcal{C}_\text{oa}$: concepts obtained by decomposing each description into \emph{objects} and \emph{actions} (verb + object). For example, ``add some parmesan cheese and salt to pot'' yields object concepts ``parmesan cheese'' and ``salt'', and action concepts ``add parmesan cheese'' and ``add salt''. This represents the video at the coarsest granularity.
\end{enumerate}

\begin{table}[tb]
  \centering
  \caption{Representative examples of concepts from each concept set.}
  \label{tab:concept-examples}
  \footnotesize
  \begin{tabularx}{\linewidth}{p{2.8cm}X} \toprule
    Concept set & Example concepts \\ \midrule
    $\mathcal{C}_\text{raw}$ &
      add soy sauce fish sauce and salt,
      cut the sushi roll into pieces,
      cook the dough in a pan,
      heat a pan and add oil,
      place the onions in the pan,
      stir the soup,
      brush the pastry with butter \ldots \\ \midrule
    $\mathcal{C}_\text{verb}$ &
      drain a sliced potato into a bowl,
      chop a tuna,
      fry a sausage in a pan,
      sprinkle cheese on a pizza,
      knead dough well,
      whisk yogurt and water,
      serve boiled eggs \ldots \\ \midrule
    $\mathcal{C}_\text{oa}$ &
      cook a sauce,
      add vanilla,
      wrap a hot dog,
      spread sauce,
      mince ginger,
      maple syrup,
      green chilli,
      honey,
      crouton \ldots \\ \bottomrule
  \end{tabularx}
\end{table}

Let $\mathcal{A} = \bigcup_{U\in\mathcal{U}} U$ denote the union of step descriptions. $\mathcal{C}_\text{raw}$ requires no LLM call; its candidate set is simply $\hat{\mathcal{C}}_\text{raw} = \mathcal{A}$. For $\mathcal{C}_\text{verb}$ and $\mathcal{C}_\text{oa}$, we use prompts $P_\text{verb}$ and $P_\text{oa}$, respectively (full prompts are provided in \cref{sec:appendix-prompts}). Letting $G$ denote the LLM, which extracts a set of concepts $G(u, P)$ from $u \in \mathcal{A}$ using prompt $P$, the concept set for $\ell \in \{\text{verb}, \text{oa}\}$ is
\begin{equation}
  \hat{\mathcal{C}}_\ell = \bigcup_{u \in \mathcal{A}} G(u, P_\ell).
\end{equation}

To remove semantically redundant concepts, we apply similarity-based filtering to these candidate concept sets. We use the Perception Encoder (PE) \cite{bolya2025perception} as our text encoder $E_T$. For any concept pair $(c, c')$ with $c \neq c'$ both in $\hat{\mathcal{C}}_\ell$ for $\ell \in \{\text{raw}, \text{verb}, \text{oa}\}$, we compute the cosine similarity
\begin{equation}
  \mathrm{sim}(c, c') = \frac{E_T(c) \cdot E_T(c')}{\| E_T(c) \|_2 \, \| E_T(c') \|_2}.
\end{equation}
We compute the similarity over all concept pairs for $\hat{\mathcal{C}}_\ell$ and discard one member of any pair satisfying
\begin{equation}
  \mathrm{sim}(c, c') > \tau,
\end{equation}
with threshold $\tau$. Applying this filtering to $\hat{\mathcal{C}}_\ell$ yields the non-redundant concept sets $\mathcal{C}_\text{raw}$, $\mathcal{C}_\text{verb}$, and $\mathcal{C}_\text{oa}$, which are used throughout the remaining steps of our pipeline.

\subsection{Temporal Concept Score Computation}
\label{sec:method-scoring}
Given the concept set constructed above, we compute the similarity between each concept and the visual content in $x_t \in X$, converting $X$ into a temporal matrix $S$ of concept scores.

We extract features from both $x_t \in X $ and $c \in \mathcal{C}_\ell$ using the pretrained video encoder $E_V$ and text encoder $E_T$ of PE \cite{bolya2025perception}. The $i$-th entry $s_{ti}$ of $s_t$ is given by the cosine similarity between the embeddings of segment $x_t$ and concept $c_i$:
\begin{equation}
  s_{ti} = \frac{E_V(x_t) \cdot E_T(c_i)}{\| E_V(x_t) \|_2 \, \| E_T(c_i) \|_2}.
\end{equation}
Applying this to every segment-concept pair gives a representation of the entire video $V$ as a score matrix $S = (s_1, \dots, s_T) \in \mathbb{R}^{|\mathcal{C}_\ell| \times T}$, whose rows describe the temporal evolution of a given concept, and whose columns describe the concept activation at a given time.

\subsection{Transformer-Based Classification}
\label{sec:method-transformer}
The temporal concept score matrix $S$ captures the semantic content of each segment, but recognizing the overall context and the temporal order of events in the video requires aggregating information along the time axis. We use a Transformer, well-suited to sequence modeling, to predict the final video class from $S$.

We first apply a linear projection that compresses the concept dimension of $s_t$ to the Transformer's hidden dimension $d_\text{model}$, which keeps the computational cost manageable even as the number of concepts grows, \ie, $z_t = W_\text{in} s_t + b_\text{in}$, where $W_\text{in} \in \mathbb{R}^{d_\text{model} \times |\mathcal{C}_\ell| }$ and $b_\text{in} \in \mathbb{R}^{d_\text{model}}$. We then prepend a learnable classification token $z_\text{cls}$ that aggregates video-level information, and add a learnable positional encoding $p_t$ to inject temporal order. The resulting input sequence $Z$ is given by
\begin{equation}
  Z = (z_\text{cls}, z_1 + p_1, \dots, z_T + p_T)  \in \mathbb{R}^{d_\text{model} \times (T+1)}.
\end{equation}

The sequence $Z$ is then passed through $L$ stacked Transformer blocks, each composed of multi-head self-attention and a feed-forward network, which progressively learn long-range dependencies between segments and the temporal evolution of concepts. This may let the model capture dynamics in $V$ that a simple temporal average of concept scores does not fully represent.

Finally, we take the $\mathrm{[CLS]}$ token output $z_\text{cls}^{(L)}$ from the last Transformer layer and apply a linear classification head followed by softmax to obtain the predicted class distribution
\begin{equation}
  \hat{y} = \mathrm{Softmax}(W_\text{out}z_\text{cls}^{(L)} + b_\text{out}),
\end{equation}
where $W_\text{out} \in \mathbb{R}^{K \times d_\text{model}}$ and $b_\text{out} \in \mathbb{R}^{K}$ are the weight matrix and bias of the classification head, and $K$ is the number of classes. We train the projection layer, the Transformer, and the classification head by minimizing the cross-entropy loss between $\hat{y}$ and the ground-truth label $y$.

\subsection{Interpreting Predictions via Concept Importance}
\label{sec:method-interpretation}

For further interpretability, we quantify the importance of a concept by observing how the prediction changes when that concept is masked. Concretely, we mask the scores of concept $c_i$ with a zero baseline across all segments, \ie, $s_{ti} = 0$ for all $t = 1,\dots,T$, and measure the resulting change in the predicted score of the ground-truth class. If $c_i$ carries information that is important for the prediction, setting $s_{ti}$ to 0 should substantially lower the prediction score. Conversely, masking a concept irrelevant to the prediction should not change the score significantly.

For an input video with ground-truth class $y$, let
\begin{equation}
  f_y(S) = p(y \mid S)
\end{equation}
denote the prediction score of $y$ given the full concept score matrix $S$. Let $S_{(c_i)}$ denote the matrix obtained from $S$ by setting $s_{ti}=0$ for all $t$, and
\begin{equation}
  f_y(S_{(c_i)}) = p(y \mid S_{(c_i)})
\end{equation}
the prediction score of $y$ under this masked input. We define the importance of concept $c_i$ as the resulting drop in probability,
\begin{equation}
  q_i = f_y(S) - f_y(S_{(c_i)}).
\end{equation}
A larger positive value of $q_i$ indicates a greater drop in prediction score when $c_i$ is masked, and hence a concept that supports the prediction of $y$. Conversely, a negative $q_i$ indicates that masking $c_i$ raises the score of $y$, \ie, that $c_i$ works against $y$. We rank concepts by $|q_i|$ to identify those with the largest influence on the score of $y$ (\cref{sec:experiments}).

Because $q_i$ is computed by masking one concept at a time, it does not separate interactions among concepts mediated by the Transformer's self-attention mechanism. The joint effect of masking two or more concepts simultaneously may differ from the sum of their individual $q_i$ values. $q_i$ should therefore be understood as the effect of masking each concept while holding the others fixed, rather than as a strict causal attribution.

\section{Experiments}
\label{sec:experiments}

\subsection{Experimental Setup}

\paragraph{Datasets.}
We evaluate our method on YouCook2 \cite{youcook2}, a dataset of third-person cooking videos collected from YouTube spanning 89 recipe categories. Each video is approximately 5 minutes long on average and is annotated with step boundaries (start and end times) together with an English sentence describing each step. We use these step descriptions as the source for concept extraction (\cref{sec:method-concepts}) and classify a video into one of 89 recipe categories. We use the standard YouCook2 split: 1,333 training, 457 validation, and 210 test videos. We also evaluate on ActivityNet \cite{activitynet} to cover a domain other than cooking. We use the 44 classes of the ``Household Activities'' category in the ActivityNet v1.3 taxonomy and take the per-video captions of ActivityNet Captions \cite{activitynet-captions} as the source of concepts. Because the test labels are not public, we train on the 1,818 available training videos and evaluate on the 899 available validation videos.

\paragraph{Concepts.}
We use Qwen3-8B \cite{qwen3} to construct $\mathcal{C}_\text{verb}$ and $\mathcal{C}_\text{oa}$ from the step descriptions in the training split of each dataset, using the prompt for each granularity described in \cref{sec:method-concepts}. For similarity-based filtering (\cref{sec:method-concepts}), we set the redundancy threshold to $\tau = 0.92$, chosen empirically to balance concept set size and redundancy mitigation. After similarity-based filtering, the concept sets for YouCook2 contain 8{,}376 raw descriptions, 10{,}492 verb phrases, and 8{,}973 object-and-action concepts (2{,}761 objects and 6{,}212 actions). Those for ActivityNet contain 6{,}211 raw descriptions, 7{,}908 verb phrases, and 8{,}993 object-and-action concepts (2{,}712 objects and 6{,}281 actions).

\paragraph{Implementation Details.}
We train our Transformer classifier with AdamW \cite{adamw}, a learning rate of $1\times10^{-4}$, a batch size of 32, for 50 epochs. We use PE (ViT-L/14@336px) \cite{bolya2025perception,vit} as the pretrained model for computing concept-segment scores. The Transformer encoder uses $L=3$ layers, 8 attention heads, and a hidden dimension of $d_\text{model}=512$. To build the input segment sequence $X$ from a video $V$, we sample with a 6-second window and a 3-second stride. All experiments were run on a single NVIDIA GeForce RTX 3090 GPU (24GB). We report Top-1 accuracy as our evaluation metric.

\paragraph{Baseline.}
To assess the effect of introducing conceptual representation, we compare against a baseline that does not route predictions through an intermediate conceptual representation. The baseline feeds video features extracted with the same encoder (PE's video encoder) directly into a Transformer classifier for training and prediction, allowing us to quantify the effect of the concept bottleneck on accuracy relative to using visual features directly.

\subsection{Results}
\subsubsection{Performance Evaluation.}

\begin{table}[tb]
  \centering
  \caption{Top-1 accuracy (\%) on YouCook2 and ActivityNet (Household Activities). Columns indicate the concept set ($\mathcal{C}_\text{raw}$: raw step descriptions, $\mathcal{C}_\text{verb}$: verb phrases, $\mathcal{C}_\text{oa}$: object-and-action). Results are the mean and standard deviation over 10 runs for YouCook2 and 3 runs for ActivityNet.}
  \label{tab:performance}
  \setlength{\tabcolsep}{3pt}
  \footnotesize

  \begin{tabular}{lcccccc}\toprule
     & \multicolumn{3}{c}{YouCook2} & \multicolumn{3}{c}{ActivityNet} \\
    \cmidrule(lr){2-4} \cmidrule(lr){5-7}
    Ours & $\mathcal{C}_\text{raw}$ & $\mathcal{C}_\text{verb}$ & $\mathcal{C}_\text{oa}$ & $\mathcal{C}_\text{raw}$ & $\mathcal{C}_\text{verb}$ & $\mathcal{C}_\text{oa}$ \\ \midrule
    All concepts & $72.4\pm1.1$ & $75.3\pm0.9$ & $73.7\pm1.4$ & $91.4\pm0.3$ & $89.7\pm0.2$ & $90.2\pm0.6$ \\
    Top-5  & $84.5\pm1.8$ & $85.3\pm1.5$ & $87.2\pm0.6$ & $91.8\pm0.1$ & $92.6\pm0.2$ & $93.0\pm0.3$ \\
    Top-10 & $88.6\pm0.6$ & $87.7\pm0.9$ & $88.6\pm0.7$ & $92.1\pm0.1$ & $93.1\pm0.1$ & $93.5\pm0.2$ \\
    Top-20 & $88.8\pm0.3$ & $89.3\pm0.5$ & $89.5\pm1.0$ & $93.5\pm0.5$ & $93.0\pm0.2$ & $93.3\pm0.3$ \\
    Top-40 & $89.4\pm0.5$ & $89.1\pm0.8$ & $88.4\pm0.8$ & $93.1\pm0.3$ & $93.5\pm0.2$ & $93.4\pm0.2$ \\
    \midrule
    \rowcolor{gray!15}
    Baseline & \multicolumn{3}{c}{$90.7 \pm 0.6$} & \multicolumn{3}{c}{$94.3 \pm 0.1$} \\
    \bottomrule
  \end{tabular}
\end{table}

\Cref{tab:performance} reports Top-1 accuracy on YouCook2 and ActivityNet. Rows indicate how the temporal concept scores are fed to the Transformer: ``All concepts'' uses every PE-computed concept score directly, while ``Top-$k$'' retains, for each segment, only the $k$ concepts with the highest similarity scores and zeroes out the rest.\footnote{For concept set $\mathcal{C}_\text{oa}$, the top $k$ are selected separately from the object and action sub-sets, yielding $2k$ concepts per segment in total. This is because the scores for object concepts and action concepts may not be directly comparable.} 

We first compare against the baseline. Training directly on video features without a conceptual representation reaches 90.7\% on YouCook2 and 94.3\% on ActivityNet. Our best models reach 89.5\% and 93.5\%, respectively, only 1.2 and 0.8 points below the baseline. This indicates that our method retains a level of discriminative power close to that of a black-box model while routing every prediction through a human-interpretable conceptual representation.

Next, we examine the effect of concept granularity. As \Cref{tab:performance} shows, with $k \geq 10$, accuracy is largely insensitive to which of the three concept sets is used on both datasets. We speculate that this is because PE's text encoder tends to weight the core verb and object of a sentence more heavily than modifying phrases such as prepositional phrases; as long as the central ``what happened to what'' is preserved, fine-grained grammatical detail may contribute little extra signal to the score. Consequently, even the most detailed concept set $\mathcal{C}_\text{raw}$ does not meaningfully outperform the much simpler, object- and action-centric concept set $\mathcal{C}_\text{oa}$.

Finally, we discuss the effect of Top-$k$ filtering. Across all three concept sets and both datasets, using the unfiltered ``All concepts'' setting performs worse than any Top-$k$ filtering, markedly so on YouCook2. We speculate that this is because the large majority of the several thousand concepts are unrelated to a given segment; their accumulation as noise may hinder the Transformer's ability to learn meaningful temporal patterns. Among the filtered settings, $k=5$ is consistently among the weakest, suggesting that retaining so few concepts per segment is too restrictive and discards information necessary for the prediction. Between $k=20$ and $k=40$, accuracy saturates, indicating that $k=20$ already retains most of the information relevant to the prediction. Overall, $k=20$ offers the best balance between coverage and noise suppression for this task.

\subsubsection{Interpretability.}

\begin{figure}[tb]
  \centering
  \includegraphics[width=0.95\linewidth]{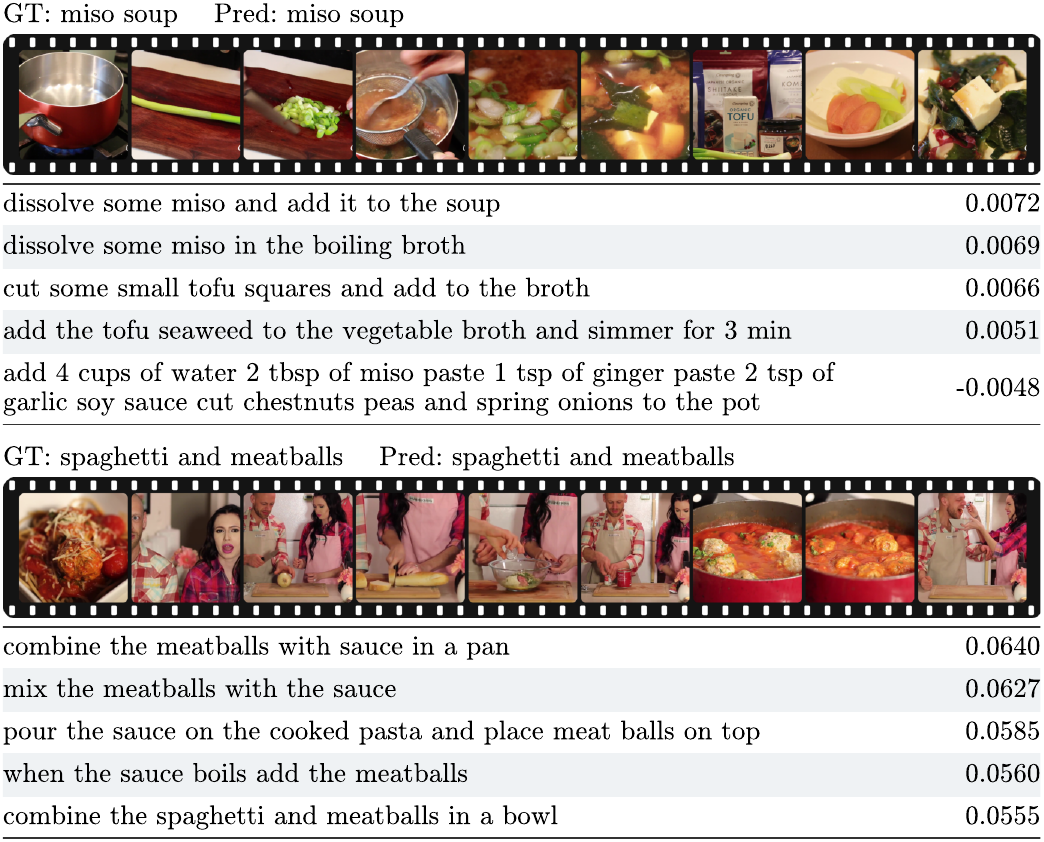}
  \caption{Visualization of the prediction rationale based on the importance score $q_i$ for two correctly classified test videos, ``miso soup'' and ``spaghetti and meatballs.''}
  \label{fig:explainable}
\end{figure}

We assess the rationale produced by our method by visualizing the importance score $q_i$ (\cref{sec:method-interpretation}) for YouCook2 test videos. \Cref{fig:explainable} shows the top-5 concepts by $|q_i|$ for videos from the ``miso soup'' and ``spaghetti and meatballs'' classes, using concept set $\mathcal{C}_\text{raw}$ with $k=20$. As defined in \cref{sec:method-interpretation}, a high-$q_i$ concept is one whose masking causes a large drop in the prediction score, and therefore represents information our model treats as central to its decision. For ``miso soup,'' the highest-ranked concept is ``dissolve some miso and add it to the soup,'' followed by ``dissolve some miso in the boiling broth''; both correspond directly to the scene in which miso is dissolved into the broth, a decisive step in the dish. Similarly, for ``spaghetti and meatballs'', concepts such as ``combine the meatballs with sauce in a pan'' and ``mix the meatballs with the sauce'' rank highly, capturing the step where the meatballs and sauce are combined, a step close to the dish's completed form.

Notably, the magnitude of the importance scores differs sharply between these two examples. The maximum score for ``spaghetti and meatballs'' is approximately $0.06$, an order of magnitude larger than the roughly $0.007$ maximum for ``miso soup''. One possible explanation is a difference in how evidence is distributed across the video. Cooking ``miso soup'' involves several similarly important steps (preparing the stock, cutting ingredients, dissolving the miso), so our model appears to combine multiple, individually modest pieces of evidence rather than relying on a single decisive concept, which lowers the importance attributed to any one concept. In contrast, ``spaghetti and meatballs'' has a more singularly decisive object and action, the meatball itself and the act of combining it with sauce, so importance concentrates on a small number of concepts, yielding a higher peak score.

\begin{figure}[tb]
  \centering
  \includegraphics[width=0.95\linewidth]{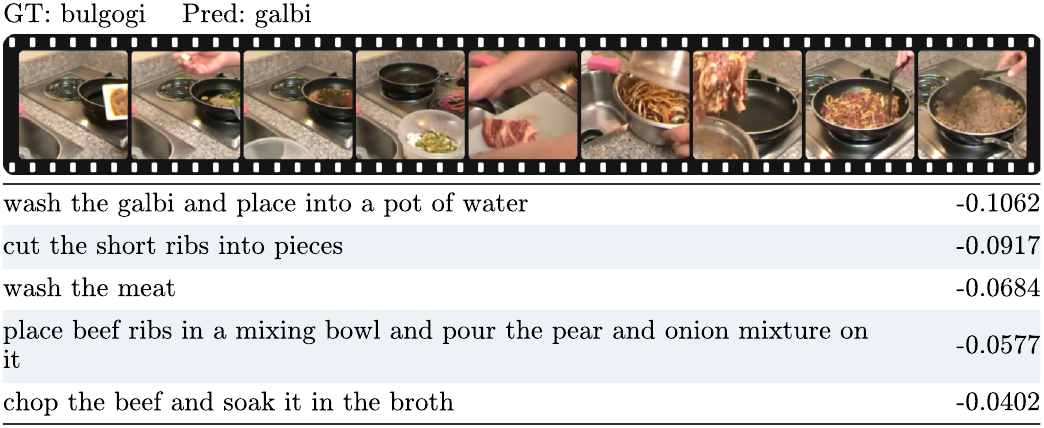}
  \caption{Visualization of the important concepts for an incorrectly classified test video. Concepts are ranked by $|q_i|$ with respect to the ground-truth class.}
  \label{fig:misclassified}
\end{figure}

Our importance scores can also help explain incorrect predictions. \Cref{fig:misclassified} shows a video of \emph{bulgogi} misclassified as \emph{galbi}. All five highest-ranked concepts have negative $q_i$, meaning that masking any of them raises the score of \emph{bulgogi}. These concepts describe galbi and short ribs (\eg, ``wash the galbi and place into a pot of water'' and ``cut the short ribs into pieces''), which suggests that the model matched the beef-handling scenes in this video to galbi-specific steps. Bulgogi and galbi are both Korean beef dishes with similar ingredients and preparation steps, which may make such confusion likely. The importance scores thus reveal which concepts kept the model from predicting the correct class, information that could remain hidden inside a black-box model.

\subsection{Ablation Study}
\subsubsection{Effect of Temporal Modeling.}

\begin{table}[tb]
  \centering
  \caption{Effect of temporal modeling (YouCook2, concept set $\mathcal{C}_\text{oa}$, Top-20).}
  \label{tab:ablation-transformer}
  \begin{tabular}{lc}\toprule
    Model variant & Top-1 Acc. (\%) \\ \midrule
    w/ Transformer  & $89.5\pm1.0$ \\
    w/o Transformer & $86.0\pm0.4$ \\ \bottomrule
  \end{tabular}
\end{table}

To verify the contribution of the Transformer-based temporal modeling at the core of our method, we compare against a variant that discards the temporal order of concept scores. This variant replaces the Transformer with average pooling of the temporal concept score matrix $S \in \mathbb{R}^{|\mathcal{C}_\ell| \times T}$ over time, yielding a single global concept score vector per video, which is then classified with a two-layer MLP. This variant still reflects how strongly each concept appears over the video, but ignores the order in which concepts occur.

\Cref{tab:ablation-transformer} compares Top-1 accuracy for concept set $\mathcal{C}_\text{oa}$ with Top-20 filtering. Modeling the temporal evolution of concepts with a Transformer improves accuracy by roughly 3 points over the pooled variant, suggesting that temporal information is useful for video recognition in our framework. This relatively modest gap is likely attributable to the nature of YouCook2. Recipe categories can often be predicted from the static presence of particular ingredients or utensils alone, so even the pooled variant retains a fairly strong signal. The Transformer, which can also model the order in which concepts occur, further improves on this signal.

\subsubsection{Effect of the Concept Source.}

\begin{table}[tb]
  \centering
  \caption{Effect of the concept source (YouCook2, Top-20).}
  \label{tab:llm-concepts}
  \begin{tabular}{lc}\toprule
    Concept set & Top-1 Acc. (\%) \\ \midrule
    $\mathcal{C}_\text{oa}$ (dataset-specific)               & $89.5\pm1.0$ \\
    LLM-generated (dataset-agnostic)        & $86.6\pm0.9$ \\
    \bottomrule
  \end{tabular}
\end{table}

The experiments above use concept sets built from dataset-specific descriptions. In a real deployment, however, such descriptive text may not always be available. We therefore evaluate a concept set generated purely from an LLM's general knowledge, without access to the associated descriptions $\mathcal{U}$. Using an LLM and the instruction ``please think of over 1{,}000 actions or ingredients related to cooking,'' we generated 1{,}895 generic cooking-related concepts without referencing YouCook2 in any way, and trained and evaluated our best-performing configuration (Top-20) on this concept set.

\Cref{tab:llm-concepts} compares this dataset-agnostic concept set with the dataset-specific object-and-action set $\mathcal{C}_\text{oa}$. Accuracy drops from 89.5\% to 86.6\%, a decrease of about 3 points. Note that the two sets also differ in size (8{,}973 vs.\ 1{,}895 concepts), so this gap may reflect both the source and the number of concepts. Given that this concept set was generated without any access to YouCook2, the drop is modest. This result suggests that our method may not rely heavily on dataset-specific textual knowledge. Even with only generic culinary knowledge and no human-written, per-video descriptions, our framework achieves over 86\% accuracy, indicating some degree of generalization to settings where detailed human annotations are unavailable.

\section{Conclusion}
\label{sec:conclusion}

We presented a Transformer-based procedural video recognition framework that uses human-interpretable concepts as an intermediate representation, addressing the black-box nature of video recognition models by routing every prediction through a temporal sequence of concept scores.

Our experiments yield two main findings. First, imposing a conceptual representation on the prediction process, when combined with Transformer-based temporal modeling, preserves accuracy close to that of a conventional black-box model. Second, with $k \geq 10$, the choice of concept granularity has only a limited effect on accuracy on both YouCook2 and ActivityNet. Detailed natural-language descriptions and much simpler object-and-action concepts achieve comparably high accuracy. This indicates that our framework may not depend heavily on fine-grained linguistic structure to recognize complex events, capturing instead the temporal evolution of the core semantic units of a video, a result that could afford flexibility in concept design.

Several limitations remain. The framework depends heavily on the VLM that aligns video segments with textual concepts. Its accuracy and explanations therefore inherit the VLM's errors and biases. The importance score $q_i$ masks one concept at a time with a zero baseline and does not separate interactions among concepts, so it does not guarantee a causal interpretation. Finally, while generic LLM-generated concepts achieve competitive accuracy, domain-specific text still provides a measurable benefit, suggesting that performance may degrade in domains where such descriptions are unavailable.

Future work includes refining the predicted rationale by grounding concepts in specific spatial regions or segments of the video, rather than the video as a whole; evaluating how convincing the presented concepts are to human users through user studies; and moving beyond a fixed concept set toward a framework that dynamically generates or selects the concepts most relevant to a given video's content. We also plan to evaluate our framework on other types of videos, such as short single-action clips.

\subsubsection*{Acknowledgements.}
This work was partly supported by JSPS KAKENHI Grant Number JP23H00497 and JST FOREST Grant Number JPMJFR216O.

%
%
\bibliographystyle{splncs04}
\bibliography{main}

\begin{thebibliography}{10}
\providecommand{\url}[1]{\texttt{#1}}
\providecommand{\urlprefix}{URL }
\providecommand{\doi}[1]{https://doi.org/#1}

\bibitem{bau2017network}
Bau, D., Zhou, B., Khosla, A., Oliva, A., Torralba, A.: Network dissection:
  Quantifying interpretability of deep visual representations. In: Proc. IEEE
  Conference on Computer Vision and Pattern Recognition (CVPR). pp. 6541--6549
  (2017)

\bibitem{timesformer}
Bertasius, G., Wang, H., Torresani, L.: Is space-time attention all you need
  for video understanding? In: Proc. International Conference on Machine
  Learning (ICML). vol.~139, pp. 813--824 (2021)

\bibitem{bolya2025perception}
Bolya, D., Huang, P.Y., Sun, P., Cho, J.H., Madotto, A., Wei, C., Ma, T., Zhi,
  J., Rajasegaran, J., Rasheed, H., et~al.: Perception encoder: The best visual
  embeddings are not at the output of the network. arXiv preprint
  arXiv:2504.13181  (2025)

\bibitem{legrad2025}
Bousselham, W., Boggust, A., Chaybouti, S., Strobelt, H., Kuehne, H.: {LeGrad}:
  An explainability method for vision transformers via feature formation
  sensitivity. In: Proc. IEEE International Conference on Computer Vision
  (ICCV). pp. 20336--20345 (2025)

\bibitem{activitynet}
Caba~Heilbron, F., Escorcia, V., Ghanem, B., Niebles, J.C.: {ActivityNet}: A
  large-scale video benchmark for human activity understanding. In: Proc. IEEE
  Conference on Computer Vision and Pattern Recognition (CVPR). pp. 961--970
  (2015)

\bibitem{i3d}
Carreira, J., Zisserman, A.: Quo vadis, action recognition? {A} new model and
  the {Kinetics} dataset. In: Proc. IEEE Conference on Computer Vision and
  Pattern Recognition (CVPR) (2017)

\bibitem{grad-cam++}
Chattopadhay, A., Sarkar, A., Howlader, P., Balasubramanian, V.N.:
  {Grad-CAM++}: Generalized gradient-based visual explanations for deep
  convolutional networks. In: Proc. IEEE Winter Conference on Applications of
  Computer Vision (WACV). pp. 839--847. IEEE (2018)

\bibitem{interactive-cbm}
Chauhan, K., Tiwari, R., Freyberg, J., Shenoy, P., Dvijotham, K.: Interactive
  concept bottleneck models. In: Proc. AAAI Conference on Artificial
  Intelligence. vol.~37, pp. 5948--5955 (2023)

\bibitem{openclip-scaling}
Cherti, M., Beaumont, R., Wightman, R., Wortsman, M., Ilharco, G., Gordon, C.,
  Schuhmann, C., Schmidt, L., Jitsev, J.: Reproducible scaling laws for
  contrastive language-image learning. In: Proc. IEEE Conference on Computer
  Vision and Pattern Recognition (CVPR) (2023)

\bibitem{vitshapley2023}
Covert, I., Kim, C., Lee, S.I.: Learning to estimate {Shapley} values with
  vision transformers. In: Proc. International Conference on Learning
  Representations (ICLR) (2023)

\bibitem{vit}
Dosovitskiy, A., Beyer, L., Kolesnikov, A., Weissenborn, D., Zhai, X.,
  Unterthiner, T., Dehghani, M., Minderer, M., Heigold, G., Gelly, S.,
  Uszkoreit, J., Houlsby, N.: An image is worth 16x16 words: Transformers for
  image recognition at scale. In: Proc. International Conference on Learning
  Representations (ICLR) (2021)

\bibitem{cem}
Espinosa~Zarlenga, M., Barbiero, P., Ciravegna, G., Marra, G., Giannini, F.,
  Diligenti, M., Shams, Z., Precioso, F., Melacci, S., Weller, A., et~al.:
  Concept embedding models: Beyond the accuracy-explainability trade-off.
  Advances in Neural Information Processing Systems (NeurIPS)  \textbf{35},
  21400--21413 (2022)

\bibitem{slowfast}
Feichtenhofer, C., Fan, H., Malik, J., He, K.: {SlowFast} networks for video
  recognition. In: Proc. IEEE International Conference on Computer Vision
  (ICCV) (2019)

\bibitem{revex2024}
Gaya-Morey, F.X., Buades-Rubio, J.M., MacKenzie, I.S., Manresa-Yee, C.:
  {REVEX}: A unified framework for removal-based explainable artificial
  intelligence in video. arXiv preprint arXiv:2401.11796  (2024)

\bibitem{hipe}
Gulshad, S., Long, T., van Noord, N.: Hierarchical explanations for video
  action recognition. In: Proc. IEEE Conference on Computer Vision and Pattern
  Recognition Workshops (CVPRW) (2023)

\bibitem{swagv2022}
Hartley, T., Sidorov, K., Willis, C., Marshall, D.: {SWAG-V}: Explanations for
  video using superpixels weighted by average gradients. In: Proc. IEEE/CVF
  Winter Conference on Applications of Computer Vision (WACV). pp. 604--613
  (2022)

\bibitem{codex2022}
Jeyakumar, J.V., Dickens, L., Garcia, L., Cheng, Y.H., Echavarria, D.R., Noor,
  J., Russo, A., Kaplan, L., Blasch, E., Srivastava, M.: Automatic concept
  extraction for concept bottleneck-based video classification. arXiv preprint
  arXiv:2206.10129  (2022)

\bibitem{align}
Jia, C., Yang, Y., Xia, Y., Chen, Y.T., Parekh, Z., Pham, H., Le, Q.V., Sung,
  Y., Li, Z., Duerig, T.: Scaling up visual and vision-language representation
  learning with noisy text supervision. In: Proc. International Conference on
  Machine Learning (ICML) (2021)

\bibitem{layer-cam}
Jiang, P.T., Zhang, C.B., Hou, Q., Cheng, M.M., Wei, Y.: {LayerCAM}: Exploring
  hierarchical class activation maps for localization. IEEE Transactions on
  Image Processing  \textbf{30},  5875--5888 (2021)

\bibitem{tcav}
Kim, B., Wattenberg, M., Gilmer, J., Cai, C., Wexler, J., Viegas, F., Sayres,
  R.: Interpretability beyond feature attribution: Quantitative testing with
  concept activation vectors ({TCAV}). In: Proc. International Conference on
  Machine Learning (ICML). pp. 2668--2677. PMLR (2018)

\bibitem{probabilistic-cbm}
Kim, E., Jung, D., Park, S., Kim, S., Yoon, S.: Probabilistic concept
  bottleneck models. In: Proc. International Conference on Machine Learning
  (ICML) (2023)

\bibitem{knab2025motif}
Knab, P., Marton, S., Schubert, P.J., Guggiana, D., Bartelt, C.: Concepts in
  motion: Temporal concept bottleneck model for interpretable video
  classification. In: Proc. International Conference on Machine Learning (ICML)
  Workshop on Compositional Learning (CompLearn) (2026)

\bibitem{cbm}
Koh, P.W., Nguyen, T., Tang, Y.S., Mussmann, S., Pierson, E., Kim, B., Liang,
  P.: Concept bottleneck models. In: Proc. International Conference on Machine
  Learning (ICML). pp. 5338--5348. PMLR (2020)

\bibitem{video-concept-discovery}
Kowal, M., Dave, A., Ambrus, R., Gaidon, A., Derpanis, K.G., Tokmakov, P.:
  Understanding video transformers via universal concept discovery. In: Proc.
  IEEE Conference on Computer Vision and Pattern Recognition (CVPR) (2024)

\bibitem{activitynet-captions}
Krishna, R., Hata, K., Ren, F., Fei-Fei, L., Niebles, J.C.: Dense-captioning
  events in videos. In: Proc. IEEE International Conference on Computer Vision
  (ICCV). pp. 706--715 (2017)

\bibitem{dance}
Lee, J., Lee, W., Park, G.M., Kim, S.T., Choi, J.: Disentangled concepts speak
  louder than words: Explainable video action recognition. In: Advances in
  Neural Information Processing Systems (NeurIPS) (2025)

\bibitem{pcbear}
Lee, J., Lee, W., Park, G.M., Kim, S.T., Choi, J.: {PCBEAR}: Pose concept
  bottleneck for explainable action recognition. In: Proc. IEEE/CVF Conference
  on Computer Vision and Pattern Recognition Workshops (CVPRW) (2025)

\bibitem{li2021towards}
Li, Z., Wang, W., Li, Z., Huang, Y., Sato, Y.: Towards visually explaining
  video understanding networks with perturbation. In: Proc. IEEE/CVF Winter
  Conference on Applications of Computer Vision (WACV). pp. 1120--1129 (2021)

\bibitem{tsm}
Lin, J., Gan, C., Han, S.: {TSM}: Temporal shift module for efficient video
  understanding. In: Proc. IEEE International Conference on Computer Vision
  (ICCV) (2019)

\bibitem{video-swin}
Liu, Z., Ning, J., Cao, Y., Wei, Y., Zhang, Z., Lin, S., Hu, H.: Video {Swin
  Transformer}. In: Proc. IEEE Conference on Computer Vision and Pattern
  Recognition (CVPR) (2022)

\bibitem{adamw}
Loshchilov, I., Hutter, F.: Decoupled weight decay regularization. In: Proc.
  International Conference on Learning Representations (ICLR) (2019)

\bibitem{shap}
Lundberg, S.M., Lee, S.I.: A unified approach to interpreting model
  predictions. In: Advances in Neural Information Processing Systems (NeurIPS)
  (2017)

\bibitem{label-free-cbm}
Oikarinen, T., Das, S., Nguyen, L.M., Weng, T.W.: Label-free concept bottleneck
  models. In: Proc. International Conference on Learning Representations (ICLR)
  (2023)

\bibitem{rise}
Petsiuk, V., Das, A., Saenko, K.: {RISE}: Randomized input sampling for
  explanation of black-box models. In: Proc. British Machine Vision Conference
  (BMVC) (2018)

\bibitem{clip}
Radford, A., Kim, J.W., Hallacy, C., Ramesh, A., Goh, G., Agarwal, S., Sastry,
  G., Askell, A., Mishkin, P., Clark, J., et~al.: Learning transferable visual
  models from natural language supervision. In: Proc. International Conference
  on Machine Learning (ICML). pp. 8748--8763. PMLR (2021)

\bibitem{lime}
Ribeiro, M.T., Singh, S., Guestrin, C.: "{Why} should {I} trust you?"
  {Explaining} the predictions of any classifier. In: Proc. ACM SIGKDD
  International Conference on Knowledge Discovery and Data Mining (KDD). pp.
  1135--1144 (2016)

\bibitem{rudin2019stop}
Rudin, C.: Stop explaining black box machine learning models for high stakes
  decisions and use interpretable models instead. Nature Machine Intelligence
  \textbf{1},  206--215 (2019)

\bibitem{video-tcav}
Saha, A., Gupta, S., Ankireddy, S.K., Chahine, K., Ghosh, J.: Exploring
  explainability in video action recognition. In: Proc. IEEE Conference on
  Computer Vision and Pattern Recognition Workshops (CVPRW) (2024)

\bibitem{grad-cam}
Selvaraju, R.R., Cogswell, M., Das, A., Vedantam, R., Parikh, D., Batra, D.:
  {Grad-CAM}: Visual explanations from deep networks via gradient-based
  localization. In: Proc. IEEE International Conference on Computer Vision
  (ICCV). pp. 618--626 (2017)

\bibitem{saliency}
Stergiou, A., Kapidis, G., Kalliatakis, G., Chrysoulas, C., Veltkamp, R.,
  Poppe, R.: Saliency tubes: Visual explanations for spatio-temporal
  convolutions. In: Proc. IEEE International Conference on Image Processing
  (ICIP). pp. 1830--1834. IEEE (2019)

\bibitem{c3d}
Tran, D., Bourdev, L., Fergus, R., Torresani, L., Paluri, M.: Learning
  spatiotemporal features with {3D} convolutional networks. In: Proc. IEEE
  International Conference on Computer Vision (ICCV) (2015)

\bibitem{attention-is-all-you-need}
Vaswani, A., Shazeer, N., Parmar, N., Uszkoreit, J., Jones, L., Gomez, A.N.,
  Kaiser, {\L}., Polosukhin, I.: Attention is all you need. In: Advances in
  Neural Information Processing Systems (NeurIPS) (2017)

\bibitem{staa2025}
Wang, Z., Liu, Y.: {STAA}: Spatio-temporal attention attribution for real-time
  interpreting transformer-based {AI} video models. IEEE Access  \textbf{13},
  101647--101661 (2025)

\bibitem{videoclip}
Xu, H., Ghosh, G., Huang, P.Y., Okhonko, D., Aghajanyan, A., Metze, F.,
  Zettlemoyer, L., Feichtenhofer, C.: {VideoCLIP}: Contrastive pre-training for
  zero-shot video-text understanding. Proc. Conference on Empirical Methods in
  Natural Language Processing (EMNLP)  (2021)

\bibitem{qwen3}
Yang, A., Li, A., Yang, B., Zhang, B., Hui, B., Zheng, B., Yu, B., Gao, C.,
  Huang, C., Lv, C., Zheng, C., Liu, D., Zhou, F., Huang, F., Hu, F., Ge, H.,
  Wei, H., Lin, H., Tang, J., Yang, J., Tu, J., Zhang, J., Yang, J., Yang, J.,
  Zhou, J., Zhou, J., Lin, J., Dang, K., Bao, K., Yang, K., Yu, L., Deng, L.,
  Li, M., Xue, M., Li, M., Zhang, P., Wang, P., Zhu, Q., Men, R., Gao, R., Liu,
  S., Luo, S., Li, T., Tang, T., Yin, W., Ren, X., Wang, X., Zhang, X., Ren,
  X., Fan, Y., Su, Y., Zhang, Y., Zhang, Y., Wan, Y., Liu, Y., Wang, Z., Cui,
  Z., Zhang, Z., Zhou, Z., Qiu, Z.: {Qwen3} technical report (2025),
  \url{https://arxiv.org/abs/2505.09388}

\bibitem{labo}
Yang, Y., Panagopoulou, A., Zhou, S., Jin, D., Callison-Burch, C., Yatskar, M.:
  Language in a bottle: Language model guided concept bottlenecks for
  interpretable image classification. In: Proc. IEEE Conference on Computer
  Vision and Pattern Recognition (CVPR) (2023)

\bibitem{post-hoc-cbm}
Yuksekgonul, M., Wang, M., Zou, J.: Post-hoc concept bottleneck models. In:
  Proc. International Conference on Learning Representations (ICLR) (2023)

\bibitem{zeiler2014visualizing}
Zeiler, M.D., Fergus, R.: Visualizing and understanding convolutional networks.
  In: Proc. European Conference on Computer Vision (ECCV). pp. 818--833 (2014)

\bibitem{siglip}
Zhai, X., Mustafa, B., Kolesnikov, A., Beyer, L.: Sigmoid loss for language
  image pre-training. In: Proc. IEEE International Conference on Computer
  Vision (ICCV) (2023)

\bibitem{ibd}
Zhou, B., Sun, Y., Bau, D., Torralba, A.: Interpretable basis decomposition for
  visual explanation. In: Proc. European Conference on Computer Vision (ECCV).
  pp. 119--134 (2018)

\bibitem{youcook2}
Zhou, L., Xu, C., Corso, J.: Towards automatic learning of procedures from web
  instructional videos. In: Proc. AAAI Conference on Artificial Intelligence
  (AAAI). vol.~32 (2018)

\end{thebibliography}

\clearpage
\appendix
\section{Prompts for Concept Generation}
\label{sec:appendix-prompts}

Below are the full prompts used for LLM-based concept extraction in our method (\cref{sec:method-concepts}).
The concept set $\mathcal{C}_\text{raw}$ requires no LLM call, since each raw step description is used verbatim as a concept.
Prompt $P_\text{verb}$ is applied to individual step descriptions to construct the verb-phrase concept set $\mathcal{C}_\text{verb}$.
The object-and-action concept set $\mathcal{C}_\text{oa}$ is constructed from two prompts, $P_{\text{oa},\text{act}}$ and $P_{\text{oa},\text{obj}}$, which extract the action and object concepts, respectively, from each step description.
We use the same prompts for both YouCook2 and ActivityNet.
The dataset-agnostic concept set (LLM-generated) discussed in the concept-source ablation (\cref{tab:llm-concepts}) is produced instead by a single open-ended prompt without referencing any dataset text.

\begin{lstlisting}[title={Prompt $P_\text{verb}$: Verb-phrase extraction (for $\mathcal{C}_\text{verb}$)}, label=lst:pverb]
Your task is to extract key verb phrases from a given sentence and convert them into a generalized format.

# Instructions
1.  Identify the main verbs and their complete action phrases (including objects and prepositional phrases).
2.  If a subsequent action has an omitted object, you must infer it from the preceding context.
3.  Convert all plural countable nouns to their singular form.
4.  Replace definite articles (like "the", "her") with indefinite articles ("a" or "an") where appropriate. Mass nouns (like "batter", "soup") should not have an article.
5.  The output must be a JSON object with a single key "concepts", containing a list of the generalized action strings.

---
# Example 1
Sentence: "He chopped the vegetables and added them to the soup."
Answer:
{{
    "concepts":[
        "chop a vegetable",
        "add a vegetable to soup"
    ]
}}

# Example 2
Sentence: "She put the keys on the counter and took off her coat."
Answer:
{{
    "concepts":[
        "put a key on a counter",
        "take off a coat"
    ]
}}

# Example 3
Sentence: "Pour the batter into the pan and bake for 30 minutes."
Answer:
{{
    "concepts":[
        "pour batter into a pan",
        "bake batter for 30 minutes"
    ]
}}

# Example 4
Sentence: "add butter, flour and hot milk to a sauce pan and whisk"
Answer:
{{
    "concepts":[
        "add butter, flour and hot milk to a sauce pan",
        "whisk butter, flour and hot milk",
    ]
}}
---

# Task
Sentence: "{text}"
Answer:
\end{lstlisting}

\begin{lstlisting}[title={Prompt $P_{\text{oa},\text{act}}$: Action extraction (for $\mathcal{C}_\text{oa}$)}, label=lst:poa_act]
You are an expert tool for extracting action. Follow the instructions below to extract the action.

# Instructions
1.  Find the Action: Identify the main verb.
2.  Find the Direct Object: Ask "What is being [verbed]?". The answer is the direct object.
3. IGNORE ALL PREPOSITIONAL PHRASES (PPs).
    - Drop every prepositional phrase entirely -- i.e., any preposition (single- or multi-word) + NOUN/NP/gerund -- regardless of whether it encodes location, manner, means, purpose, time, etc.
        - Examples: on the tray, in a bowl, at room temperature, into the pan, with flour, by hand, for 5 minutes, after cooling, as a garnish, about 2 cups.
    - Always drop "with + NOUN" phrases (means/materials),
        - e.g., with flour / with salt and pepper / with oil / with sugar.
    - Do not treat patterns like "coat with ... / season with ... / dust with ... / spread with ..." as phrasal verbs. The verb is the head; the with-phrase is a PP and must be dropped.
    - Also drop time and purpose PPs: for 10 minutes, for serving, before baking, after cooling.
    - If a verb takes a PP complement (listen to music, look at pictures), still drop the PP; leave the verb without that object.
    - Exception (particles, not prepositions):
        - Do not drop verb particles (aka phrasal-verb particles) such as up/off/out/over/down/away/back/apart when they function as particles (e.g., pick up, turn off).
        - Heuristic: if the word is a separable particle or there is no NP immediately governed by it, keep it (e.g., pick it up, mix in). If it heads a PP with its own noun (in the bowl, on the tray), drop it.
4.  Handle Multiple Objects: If a verb has multiple objects linked by 'and' or commas (e.g., "add cheese and salt"), create a separate action for each object.
5.  Resolve Pronouns and Omitted Objects: If a pronoun (e.g., "it", "them") or a verb with a missing object refers to a direct object from a previous clause, replace the pronoun with the noun it refers to.
6.  Handle Multiple Verbs: If multiple verbs (e.g., "slice and dice") apply to the same object, create a separate action for each verb. Distribute the object to each verb (e.g., "slice and dice the ham" becomes "slice ham" and "dice ham").
7.  Normalize Output:
    - Extract only simple "verb + (direct object)" pairs (object optional if absent).
    - Lowercase objects; keep verbs as they appear.
    - Do not include any prepositional phrase in the concept string.
    - Return the concepts as a JSON array named "concepts". No explanations.
---

Example 1
Sentence: "Take the chicken out of the fridge and cut it into small pieces"
// Reasoning:
// 1. First Action: "Take the chicken out..."
//    - Verb: "take"
//    - Object: "the chicken"
//    - Ignore: "of the fridge" (location)
// 2. Second Action: "cut it..."
//    - Verb: "cut"
//    - Object: "it" -> resolves to "chicken"
//    - Ignore: "into small pieces" (manner)
// 3. Normalize & Combine: "take out a chicken", "cut a chicken"

Answer:
{{
    "concepts":[
        "take out a chicken",
        "cut a chicken"
    ]
}}

# Example 2
Sentence: "place the bread slices on top of each other and place in a hot pan"
// Reasoning:
// 1. First Action: "place the bread slices..."
//    - Verb: "place"
//    - Object: "bread slices"
//    - Ignore: "on top of each other" (manner)
// 2. Second Action: "place in a hot pan"
//    - Verb: "place"
//    - Object: None explicit
//    - Ignore: "in a hot pan" (location)
// 3. Normalize & Combine: "place a bread slice"

Answer:
{{
    "concepts":[
        "place a bread slice"
    ]
}}

# Example 3
Sentence: "Simmer for five minutes."
// Reasoning:
// 1. Action: "Simmer for five minutes."
//    - Verb: "simmer"
//    - Object: None (intransitive)
//    - Ignore: "for five minutes" (duration)
// 2. Normalize: "simmer"

Answer:
{{
    "concepts":[
        "simmer"
    ]
}}
---
IMPORTANT: Now, apply the exact same reasoning process to the following sentence. Your output must only be the raw JSON object.
Do not include prepositional phrases in the concepts.
**Extract only simple verb-object pairs; ignore all prepositional phrases.**
# Task
Sentence: "{text}"
Answer:
\end{lstlisting}

\begin{lstlisting}[title={Prompt $P_{\text{oa},\text{obj}}$: Object extraction (for $\mathcal{C}_\text{oa}$)}, label=lst:poa_obj]
Extract noun phrases from the sentence and list them, following the pattern in the example below.

#Instructions
- **Rule 1: Extract ONLY Nouns:** Target only words representing physical objects, substances, places, or specific concepts.
- **Rule 2: NO Verbs:** Absolutely do not extract any word used as a verb (an action or command), even if it can be a noun in other contexts. For example, ignore words like 'mix', 'stir'.
- **Rule 3: Return Empty if No Nouns:** If the sentence contains no valid nouns according to the rules above (e.g., it only contains verbs), the output "concepts" list must be empty.
- **Rule 4: Remove Time Expressions & Durations:**
    - Drop any tokens that denote time or duration (e.g., minute(s), hour(s), day(s), overnight, immediately, while, until, second(s), today, tomorrow, morning, evening).
- **Rule 5: Remove Pronouns & Deictics:**
    - Drop: it, them, this, that, these, those, everything, anything, something, nothing, oneself.
- **Rule 6: Remove Non-noun State/Quality/Adjective Tokens:**
    - Drop words denoting states/qualities or sides/positions when not functioning as nouns: brown, golden, high, low, medium, other side, bottom, edge, rest, warm, hot, cold, raw, cooked, tender, crispy, soft, smooth, fine.
- **Rule 7: Remove Measurement/Quantity Words; Keep Only the Underlying Noun:**
    - Drop: piece(s), slice(s), bunch, handful, cup(s), teaspoon(s), tablespoon(s), gram(s), kilogram(s), pound(s), ounce(s), liter(s), milliliter(s).
    - For "X of Y", keep only Y (semantic head).
    - Examples: "piece of butter" -> butter; "bowl of ground beef" -> ground beef.

---
# Example 1
Sentence: "place all the books and the notebook back on the shelf"
Answer:
{{
    "concepts":[
        "book",
        "notebook",
        "shelf"
    ]
}}

# Example 2
Sentence: "She bought a new coffee maker, some batteries, and a toothbrush at the department store."
Answer:
{{
    "concepts":[
        "coffee maker",
        "battery",
        "toothbrush",
        "department store"
    ]
}}

# Example 3
Sentence: "He bought a box of chocolates for her."
Answer:
{{
    "concepts":[
        "chocolate"
    ]
}}

# Example 4
Sentence: "boil and stir"
Answer:
{{
    "concepts":[

    ]
}}
---
#Important
Absolutely do not extract any word used as a verb (an action or command), even if it can be a noun in other contexts. For example, ignore words like 'mix', 'stir'.
Ignore all adjectives.
"whisk" and "mix" are verbs, so exclude them.
Your output must only be the raw JSON object.

# Task
Sentence: "{text}"
Answer:
\end{lstlisting}

\begin{lstlisting}[title={Prompt for LLM-generated (dataset-agnostic) concept set}, label=lst:llm]
Please think of over 1000 actions or ingredients related to cooking.
\end{lstlisting}

\end{document}